\documentclass{article}

\usepackage{microtype}
\usepackage{graphicx}
\usepackage{subfigure}
\usepackage{booktabs} 
\usepackage{colortbl}
\usepackage{xcolor}
\usepackage{caption}
\usepackage{hyperref}

\usepackage[accepted]{icml2024}

\usepackage{amsmath}
\usepackage{amssymb}
\usepackage{mathtools}
\usepackage{amsthm}

\usepackage[capitalize,noabbrev]{cleveref}

\theoremstyle{plain}

\theoremstyle{definition}

\theoremstyle{remark}

\usepackage[textsize=tiny]{todonotes}

\icmltitlerunning{Submission and Formatting Instructions for ICML 2024}

\begin{document}

\twocolumn[
\icmltitle{S\(^{2}\)-DMs: Skip-Step Diffusion Models}



\icmlsetsymbol{equal}{*}

\begin{icmlauthorlist}
\icmlauthor{Yixuan Wang}{equal,sch}
\icmlauthor{Shuangyin Li}{equal,sch}

\end{icmlauthorlist}

\icmlaffiliation{sch}{School of Computer Science, South China Normal University, Guangzhou, China}

\icmlcorrespondingauthor{Yixuan Wang}{yixuanwang@m.scnu.edu.cn}
\icmlcorrespondingauthor{Shuangyin Li}{shuangyinli@scnu.edu.cn}

\icmlkeywords{Machine Learning, ICML}

\vskip 0.3in
]



\printAffiliationsAndNotice{\icmlEqualContribution} 

\begin{abstract}
Diffusion models have emerged as powerful generative tools, rivaling GANs in sample quality and mirroring the likelihood scores of autoregressive models. A subset of these models, exemplified by DDIMs, exhibit an inherent asymmetry: they are trained over $T$ steps but only sample from a subset of 
 $T$ during generation.     This selective sampling approach, though optimized for speed, inadvertently misses out on vital information from the unsampled steps, leading to potential compromises in sample quality. To address this issue, we present the S\(^{2}\)-DMs, which is a new training method by using an innovative $L_{skip}$, meticulously designed to reintegrate the information omitted during the selective sampling phase. The benefits of this approach are manifold: it notably enhances sample quality, is exceptionally simple to implement, requires minimal code modifications, and is flexible enough to be compatible with various sampling algorithms.  On the CIFAR10 dataset, models trained using our algorithm showed an improvement of 3.27\% to 14.06\% over models trained with traditional methods across various sampling algorithms (DDIMs, PNDMs, DEIS) and different numbers of sampling steps (10, 20, ..., 1000). On the CELEBA dataset, the improvement ranged from 8.97\% to 27.08\%. 
\end{abstract}
\begin{figure}[t]
  \centering
  \includegraphics[width=\linewidth]{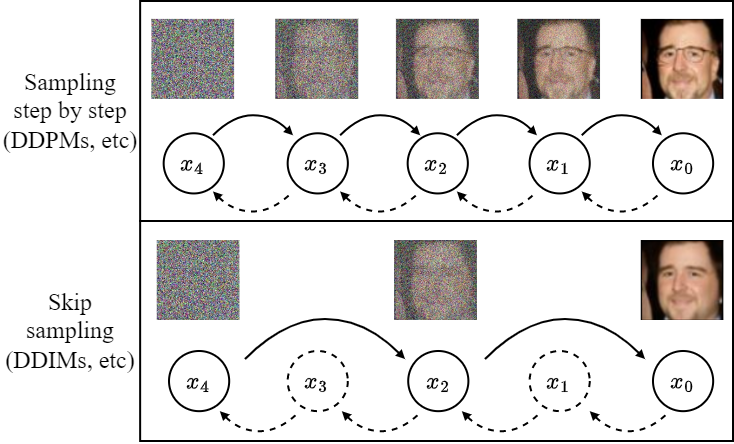}
  \caption{Overview of Different Sampling Modes in Diffusion Models. Sampling is generally divided into step-by-step sampling and skip sampling. However, all diffusion model training processes are conducted step by step, so skip sampling can lead to the loss of intermediate information, resulting in poor sample quality.}
 \label{show}
\end{figure}
\section{Introduction}
Generative models, especially deep generative models, play a foundational role in the machine learning domain \cite{karras2020analyzing}; \cite{oord2016wavenet}. Architectures like Variational Autoencoders (VAEs; \cite{kingma2013auto}) and  Autoregressive models\cite{van2016conditional};\cite{brown2020language}; \cite{salimans2017pixelcnn++}, Generative Adversarial Networks (GANs; \cite{goodfellow2014generative}; \cite{yu2017seqgan}; \cite{hjelm2017boundary};\cite{fedus2018maskgan}), and Restricted Boltzmann Machines (RBMs; \cite{hinton2012practical}) have been at the forefront. VAEs, while providing a structured probabilistic framework, occasionally yield blurry samples. GANs, acclaimed for their prowess in generating high-resolution images, can face training instabilities \cite{adler2018banach}; \cite{gulrajani2017improved}; \cite{karras2019style}. RBMs, though seminal, find themselves overshadowed by more recent architectures in scalability and performance. Against this backdrop, diffusion models, Denoising diffusion probabilistic models(DDPMs; \cite{ho2020denoising}; \cite{sohl2015deep}; \cite{song2020score}), have emerged as a compelling alternative, exhibiting unmatched capabilities in generating superior samples in diverse domains, from image synthesis to molecule design \cite{bengio2014deep}.

However, diffusion models do come with challenges. Their inherently slow sampling speed, driven by the multitude of necessary sampling steps, remains a significant concern. Recent research has honed in on this computational bottleneck, with the goal of optimizing the sampling process \cite{jolicoeur2021gotta}, \cite{nichol2021improved}. A significant breakthrough in this area is the Denoising Diffusion Implicit Models (DDIMs; \cite{song2020denoising}). DDIMs utilize a subset sampling strategy, achieving faster performance by sampling from a smaller subset instead of the entire set of steps. This method, due to its omission of certain steps, is coined "skip-step sampling." Yet, this acceleration introduces an inconsistency between training and sampling. During training, the model undergoes every step, but during sampling, some steps are selectively skipped, posing a risk of information loss. Although DDIMs, with sampling algorithm adjustments, have lessened the adverse effects of this approach compared to DDPMs, they haven't specifically addressed and optimized for the missing intermediate information. Consequently, this challenge persists, leading to the suboptimal performance of diffusion models during expedited sampling and hindering the generation of high-quality samples.

Driven by these observations, our research proposes a method of integrating skip-step sampling as an optimization target into the training process. The model trained in this way can consider and adapt to the information lost during the sampling period due to skip-step sampling. Our model not only maintains efficient generation when using other accelerated sampling algorithms but also improves the quality of the generated samples, thereby achieving a balance of efficiency and performance. Specifically, during the training process, the original loss function is retained, and at the same time, a skip-step loss is introduced, measuring the difference between the model's current step prediction and the skip-step result. This skip-step loss is combined with the original loss using a weighted mechanism. Figure \ref{show} shows an overview of the diffusion models.


Empirical results demonstrate that when the skip-step loss is incorporated into the loss objective, the S\(^2\) -DMs achieves superior performance in unconditional generation on the CIFAR10 \cite{krizhevsky2009learning} and CELEBA datasets, under the same sampling algorithms (DDIMs, PNDMs \cite{liu2022pseudo}, and DEIS \cite{zhang2022fast}), compared to models trained with the original loss objective. Importantly, our study shows that our model achieves better results than the original model at different steps, with an improvement ranging from 3.27\%-14.06\% (CIFAR10) and 8.97\%-27.08\% (CELEBA), under the same sampling algorithms. We also conducted qualitative experiments and ablation studies, along with related analyses.

The key \textbf{contributions} of this work are:

\begin{enumerate}
\item Modeling for Training-Sampling Discrepancy: To our knowledge, our study proposes the first method specifically designed to mitigate the inherent mismatch between training and sampling in diffusion models. This strategy consistently demonstrates superior performance.
\item Innovative Skip-Step Loss: We introduce a trailblazing skip-step loss, embedding the selective sampling modality directly within the training process.  This method empowers models to proactively navigate potential sampling information deficits, enhancing the quality of the samples.
\item Simplicity of Implementation: The S\(^{2}\)-DMs approach stands out not just for its efficacy but also its simplicity. With minimal code alterations required, it offers a convenient solution for both researchers and practitioners. Crucially, it's adaptable to a range of sampling algorithms.
\end{enumerate}

\section{Background}\label{2}

This study is based on DDPMs \cite{ho2020denoising} and DDIMs \cite{song2020denoising}, so a brief review is in order. DDPMs specifies a prior Markov forward diffusion process, which gradually adds noise to the data over $T$ steps. Refer to the background description of \cite{watson2021learning}. Following the notation of \cite{ho2020denoising},
\begin{equation}\label{qe}
q(x_0,...,x_T) = q(x_0)\prod_{t=1}^n q(x_t|x_{t-1}),
\end{equation} 
\begin{equation}
\begin{split}
q(x_t|x_{t-1})&=\mathcal{N}(x_t;\sqrt{\alpha_t}x_{t-1}, (1-\alpha_t)I),q(x_t|x_0)\\
&=\mathcal{N}(x_t;\sqrt{\bar\alpha_t}x_{t-1}, (1-\bar\alpha_t)I),
\end{split}
\end{equation} 
where \( q(x_0) \) represents the data distribution and \( 1-\alpha_t \) signifies the variance of the Gaussian noise added at step \( t \). For each \( t \), we have \( \alpha_t = 1 - \beta_t \) and  \( \bar{\alpha}_t = \prod_{s=1}^{t} \alpha_s \). To facilitate the transformation of noise back into data, DDPMs are trained to invert equation \ref{qe} with a model \( p_{\theta}(x_{t-1}|x_t) \). This model is trained by optimizing a (possibly reweighted) evidence lower bound (ELBO),
\begin{equation}
\begin{split}
&E_q[D_{KL}[q(x_T|x_0)||p(x_T)]+\\
&\sum_{t=2}^T D_{KL}[q(x_{t-1}|x_t,x_0)||p_\theta(x_{t-1}|x_t)]-\log{p_\theta(x_0|x_1)}],
\end{split}
\end{equation} 
\begin{equation}
\begin{split}
    &q(x_{t-1}|x_t, x_0)=\mathcal{N}(x_{t-1}; \Tilde{\mu}_t(x_t, x_0), \Tilde{\beta}_t(x_t)I) \\
    &=\mathcal{N}(x_{t-1};\frac{\sqrt{\bar{\alpha}_{t-1}}\beta_t}{1-\bar{\alpha}_t}x_0+\frac{\sqrt{\alpha_t}(1-\bar{\alpha}_{t-1})}{1-\bar{\alpha}_t}, \frac{1-\bar{\alpha}_{t-1}}{1-\bar{\alpha}_t}\beta_t I).
\end{split}
\end{equation}

DDPMs explicitly select the model for parameterization as
\begin{equation}
\begin{split}
&p_\theta(x_{t-1}|x_t)=q(x_{t-1}|x_t,\frac{1}{\sqrt{\bar{a}_t}}(x_t-\sqrt{1-\bar{a}_t}\epsilon_\theta(x_t,t)))\\
&=\mathcal{N}(x_{t-1};\frac{1}{\sqrt{\alpha_t}}(x_t-\frac{\beta_t}{\sqrt{1-\bar{a}_t}}\epsilon_\theta(x_t,t)),\frac{1-\bar{a}_{t-1}}{1-\bar{a}_{t}}\beta_tI).
  \end{split}
\end{equation} 
 In this framework, optimizing the ELBO corresponds to minimizing denoising score matching goals as explained by \cite{vincent2011connection}. \cite{song2020denoising} introduced the DDIMs concept, a set of ELBOs complemented by forward diffusion processes and sampling mechanisms. These ELBOs, having similar marginals as DDPMs, offer flexibility in determining posterior variances \cite{chen2020wavegrad}. \cite{song2020denoising} emphasized crafting alternative ELBOs using a subset of original timesteps \( S \subset \{1, ..., T\} \) with consistent marginals. This results in \( q_S(x_t|x_0) = q(x_t|x_0) \) for every \( t \) in \( S \), permitting faster sampling processes compatible with pre-trained models by integrating new timesteps. Their work also suggests the feasibility of creating a vast range of non-Markovian processes, denoted as \( \{q_\sigma : \sigma \in [0, 1]^{T -1}\} \), with each \( q_\sigma \) maintaining marginals aligned with the original progression,
\begin{equation}
q_\sigma(x_0,...,x_t)=q(x_0)q(x_T|x_0)\prod_{t=1}^{T-1}q_\sigma(x_t|x_{t+1},x_0),
\end{equation} 
and where the posteriors are defined as
\begin{equation}
\begin{split}
q_\sigma(x_{t-1}|x_t)&=\mathcal{N}(x_{t-1};\sqrt{\bar{\alpha}_{t-1}}\left(\frac{x_t-\sqrt{1-\bar{\alpha}_t}\epsilon_\theta(x_t,t)}{\sqrt{\bar{\alpha}_t}}\right)\\
&+ \sqrt{1-\bar{\alpha}_{t-1}-\beta^2_t}\epsilon_\theta(x_t, t),\beta_t^2I).
\end{split}
\end{equation} 
In their research, \cite{song2020denoising} observed that the special case of employing all-zero variances, termed as DDIMs(\( \eta = 0 \)), persistently enhances the quality of samples in the short-step domain. When amalgamated with an apt choice of timesteps for assessing the modeled score function, known as strides, DDIMs(\( \eta = 0 \)) sets a new benchmark in the realm of few-step diffusion model sampling, especially with minimal inference step allocations. A pivotal advancement we bring is the enhancement of sample quality by introducing skip information (i.e., the aforementioned subset) during the training phase, ultimately establishing a novel diffusion model training paradigm. For a more comprehensive discussion on the S$^2$-DMs, please refer to Section \ref{3}.

\section{Skip-Step Diffusion Models}\label{3}



Acceleration approaches often employ skip-step sampling as a strategy for acceleration. However, this approach inherently introduces non-smooth denoising, leading to a potential decline in performance. This observation prompted us to re-evaluate the entire training and sampling workflow. Intriguingly, we identified an asymmetry between the training and sampling phases: the former proceeds in single steps, while the latter uses skip-step.

To improve the quality of skip-step sampling, we devised a novel yet straightforward objective function for the training phase. By incorporating the skip-step objective into our original loss function (Section \ref{3.2}), we achieved a symmetrical training effect. Ultimately,  the performance of the model trained by this method met our expectations. (Section \ref{Experiments}).

\subsection{Asymmetry in Accelerated Sampling}

Due to the slow sampling speed of diffusion models, extensive research has been conducted on acceleration algorithms for these models, with DDIMs being the most prominent. In the original paper, it was stated that a subset of the full $T$ steps of the diffusion model was selected. This subset forms an increasing sequence and is considerably shorter than the original $T$ steps, leading to a significant acceleration in the sampling process. Specifically, in its implementation, not all $T$ steps are sampled. Instead, 50-step and 100-step samplings are more prevalent, which are 10-20 times faster than the original 1000-step sampling. For instance, in the 100-step sampling, the model samples at every 10th step, maintaining equal intervals between each sample.

Clearly, there's an asymmetry between the behavior during training and sampling. During training, the model is trained across all diffusion steps. In contrast, during sampling, it samples only a subset of these steps using skip-step sampling. Consequently, information from intermediate steps is overlooked, inevitably leading to a decline in model performance. We term this the ``asymmetric diffusion model."

In the subsequent sections, we will present a technique to integrate skip-step information during the training phase. This approach ensures that the trained model is more attuned to the skip-step sampling process, culminating in what we call the ``Skip-Step Diffusion Models."

\subsection{Training with Skip-Step Loss} \label{3.2}
We aim to introduce a novel skip-step loss function built upon the original one.   The standard optimization function for diffusion models was presented in the DDPMs and subsequent diffusion models predominantly utilize this foundational loss function,
\begin{equation}
    L_{t-1}=\mathbb{E}_q\Bigg[\frac{1}{2\sigma^2_t}||\tilde\mu_t(x_t, x_0)-\mu_\theta(x_t, t)||^2\Bigg]+C,
\end{equation}
we may choose the parameterization
\begin{equation}
L_0=\mathbb{E}_{t, x_0, \epsilon}|| \epsilon - \epsilon_\theta(\sqrt{\alpha_t}x_0+\sqrt{1-\alpha_t}\epsilon, t)||^2.
\end{equation} 

Initially, we assume that sampling is conducted every 10 steps, which aligns with the commonly used DDIMs setting. This configuration allows us to reduce the sampling from the original 1000 steps to just 100 steps. Here, the skip-step setting corresponds to the sampling time, denoted as $skip=10$ (Subsequent experiments will explore the model performance with various skip values). We will now introduce skip-step information, and for this purpose, we define $\alpha_{skip}^t = \alpha_t\cdot\alpha_{t+1}\dots\alpha_{t+9}$.

Then the role of \( L_0 \) is to enable the model to learn the information at each step, which corresponds to \( p_\theta(x_{t-1}|x_t) \). However, during the training phase, the corresponding sampling step is \( q(x_{t+skip}|x_{t-1}) \). Hence, we consider it as a new skip-step loss function. During the training process of the model, we aim to make $p_\theta(x_{t-1}|x_{t+skip})$ as close as possible to $q(x_t|x_{t-1})$. By doing so, when sampling with skips, the model can produce outputs that are closely aligned with the corresponding positions, thereby enhancing the quality of the output. Their formulations are as follows:
 \begin{equation}
 \begin{split}
\mu_\theta(x_{t+skip}, t+skip)&=\frac{1}{\sqrt{\alpha^{t-1}_{skip}}}(x_{t+skip}\\
&- \sqrt{1-\alpha^{t-1}_{skip}}\epsilon_\theta(x_{t+skip}, t+skip)),
 \end{split}
\end{equation} 
To achieve our desired outcome, we minimize the squared difference between our targeted optimization goal and the original target, thereby effectively reducing the discrepancy between them,
\begin{equation}
    L_{t-1}=\mathbb{E}_q\Bigg[\frac{1}{2\sigma^2_t}||\tilde\mu_t(x_t, x_0)-\mu_\theta(x_{t+skip}, t+skip)||^2\Bigg]+C,
\end{equation}


After simplification following re-parameterization, we are able to derive our final loss function $L_{skip}$,
\begin{align}
\begin{split}
L_{skip}&=  \mathbb{E}_{t+skip, x_0, \epsilon}\bigg|\bigg| \epsilon - \frac{\sqrt{1-\alpha_{skip}^{t-1}}}{\sqrt{\alpha_{skip}^{t-1}}} \epsilon_\theta ( \sqrt{\alpha_{t+skip}} x_0\\
&+ \sqrt{1-\alpha_{t+skip}} \epsilon, t+skip )\bigg|\bigg|^2.
\end{split}
\end{align}


Owing to the adjustment of certain weight values, it becomes crucial to incorporate a suitable weight in the following sections to guarantee the effective functioning of the system. At present, the stability of this loss function's value is maintained, effectively averting any system collapse. This stability is evidenced by the results detailed in the experimental section (see the experimental in section \ref{Experiments}).

\begin{figure}[t]
  \centering
  \includegraphics[width=\linewidth]{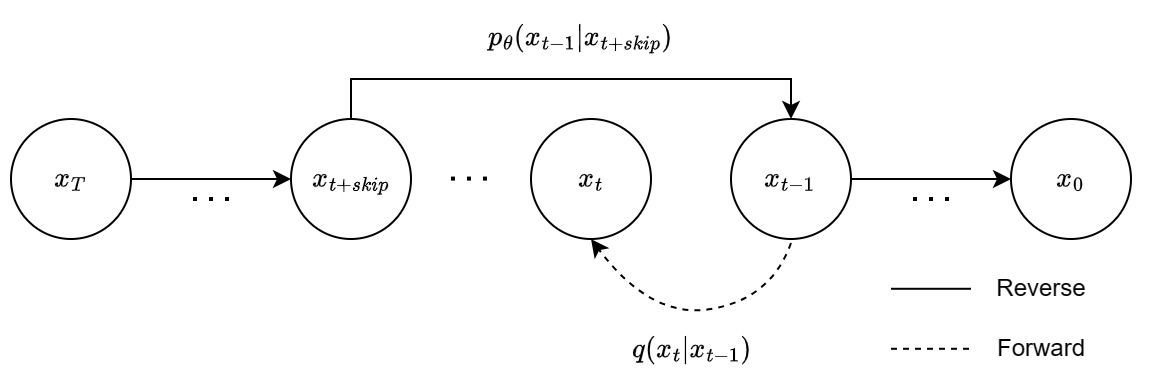}
  \caption{The directed graphical model of the S\(^2\)-DMs.
  }
  \label{pic:model}

\end{figure}
\begin{figure*}[t]
  \centering
  \includegraphics[width=0.9\linewidth]{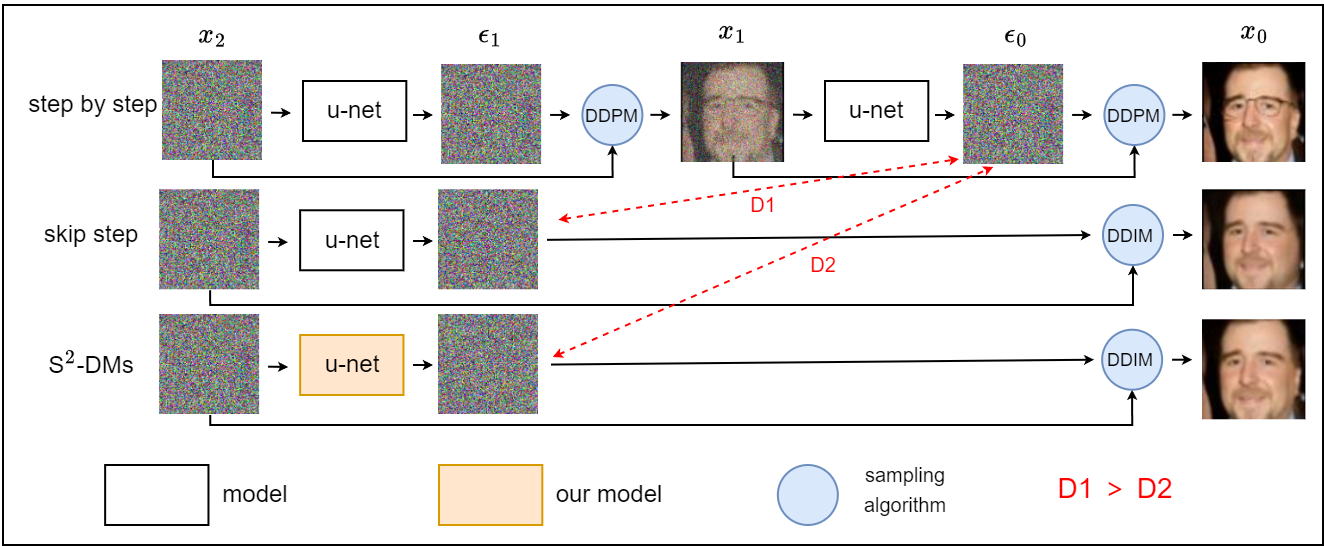}
  \caption{
    Overview of reverse process, comparing the step by step models (DDPMs, etc) and the skip models (DDIMs, PNDMs, DEIs, etc), and our S$^2$-DM trained method. The step by step models predict noise at each step for sampling, whereas the skip models accelerate the process by using current noise predictions for next steps later. Our model trained by S$^2$-DM method integrates skip-step information during training and predicts current noise like the skip models. However, the noise predicted by our model more closely resembles what the stpe by step models would predict next steps ahead, reducing the gap and enhancing sampling quality.
  }
  \label{pic:models}

\end{figure*}
\subsection{Loss scaling}

Traditionally, the training of diffusion models is centered around a singular training objective, often denoted as $L_{0}$. This conventional approach focuses on a specific set of parameters to optimize the model's performance. However, our innovative S²-DMs paradigm introduces an additional training objective, $L_{skip}$, which represents a significant shift from the traditional methodology. The incorporation of $L_{skip}$ is designed to specifically address and mitigate the challenges associated with skip-step diffusion processes.

To effectively integrate these two objectives, we propose a balanced weighting approach: $L=\tau L_0 + (1-\tau)L_{skip}$. Here, $\tau$ serves as a tuning parameter, enabling us to adjust the relative influence of the original training objective ($L_0$) and the new skip-step objective ($L_skip$). This balanced formula is crucial in harmonizing the traditional and novel aspects of our model, ensuring that neither is disproportionately emphasized at the expense of the other.

Furthermore, from a conceptual standpoint, the form of the training objective in S$^2$-DMs bears resemblance to a regularization term. This similarity is not merely coincidental but stems from a deliberate design choice. The primary motivation behind this design is to provide informational compensation for the skip-step diffusion model. By doing so, we aim to guide and constrain the model's trajectory more effectively. This approach is somewhat analogous to the principle of regularization in machine learning, where additional information is provided to prevent overfitting and to enhance the model's generalization capabilities. In the context of S$^2$-DMs, this 'regularization-like' approach aids in refining the model's performance, particularly in handling the complexities introduced by the skip-step diffusion process.

\subsection{Training and Sampling}
Figure \ref{pic:models} highlights our method's core. By integrating skip-step data during training, the model gains a broader perspective, enhancing sampling performance. While the model may lean towards symmetry, it's not a prerequisite for optimal performance. Peak efficacy is seen when nearing a symmetric form, termed "coordinated". This aligns with the model utilizing both current and post-skip data for improved predictions. The model remains flexible, not restricted by the $skip$ parameter, allowing diverse step sampling.

Incorporating $L_{skip}$ during training doesn't introduce a new sampling method but modifies the diffusion model's traditional training approach. In essence, models following the original diffusion training can benefit from our method. Our tests, using different sampling algorithms on identically trained models, consistently matched our predictions (see Experiment \ref{Experiments}).
\begin{minipage}{\linewidth}

\begin{algorithm}[H]
\caption{S\(^{2}\)-DMs Training process.}
\label{alg:algorithm1}
\begin{algorithmic}[1] 
\REPEAT
\STATE $x_0\sim q(x_0)$;
\STATE $t \sim Uniform({1,...,T})$;
\STATE $\epsilon \sim \mathcal{N}(0,I)$;
\STATE $L_0 = \nabla_\theta ||\epsilon - \epsilon_\theta||^2$;
\STATE $L_{skip} = \nabla_\theta ||\epsilon - \frac{\sqrt{1-\alpha_{skip}^{t-1}}}{\sqrt{\alpha_{skip}^{t-1}}} \epsilon_\theta||^2$;
\STATE $\begin{array}{l}
\text{Take gradient descent step on:}\\(1-\tau) \cdot L_0 + \tau \cdot L_{skip};
\end{array}$
\UNTIL{convergence is achieved}
\end{algorithmic}
\end{algorithm}

\end{minipage}
\begin{minipage}{\linewidth}

\begin{algorithm}[H]
\caption{S\(^{2}\)-DMs sampling with DDIMs.}
\label{alg:algorithm2}
\begin{algorithmic}[1] 
\vspace{0.15em}
\REPEAT
\STATE \( x_T \sim \mathcal{N}(0,1) \);
\FOR{\( t=T,...,1 \)}
\STATE  $if\ t > 0:\ \sigma \sim \mathcal{N}(0,I)\ $ \\$ else:\ \sigma = 0
 $ ;
\STATE $\begin{array}{l}
x_{t-1} = \sqrt{\bar{\alpha}_{t-1}}\left(\frac{x_t-\sqrt{1-\bar{\alpha}_t}\epsilon_\theta}{\sqrt{\bar{\alpha}_t}}\right)
\\ \quad + \sqrt{1-\bar{\alpha}_{t-1}-\sigma^2}\epsilon_\theta + \sigma^2\epsilon\ ;
\end{array}$
\ENDFOR
\vspace{0.15em}

\UNTIL{convergence is achieved}
\end{algorithmic}
\end{algorithm}
\end{minipage}

In Algorithm \ref{alg:algorithm1} and \ref{alg:algorithm2}, we illustrate the training and sampling procedures of the S$^2$-DMs. The training process, compared to the standard diffusion models, only involves an additional computation of $L_{skip}$ . This computation is straightforward. In our code repository, one can see that only a few lines of the entire code were modified to achieve all changes, making it easy to implement and facilitating follow-up by other researchers. The sampling procedure follows the standard DDIMs sampling. As skip-step information was incorporated during the training, no modifications are required in the sampling process. This allows for the generation of higher-quality samples, making it user-friendly.

\begin{table*}[th]
    \centering
    \caption{FID scores for the S$^2$-DMs against baseline methods trained on CIFAR10(32x32) with the $L_{skip}$. CIFAR10 and CelebA were training for 600K/400K iterations. For the S$^2$-DMs(DDIMs), S$^2$-DMs(PNDMs) and S$^2$-DMs(DEIS), we set $skip=10$, while other parameters remained consistent, and it was trained for 600K/400K iterations.
    }
    \label{table:cifar10}
    \begin{tabular}{ccccccc}
        \hline
        \textbf{Models} \textbackslash \, \# samplesteps S & \textbf{10} & \textbf{20} & \textbf{50} & \textbf{100} & \textbf{200} & \textbf{1000} \\
        \hline \hline
        \noalign{\vskip 3pt}
        DDIMs & 12.98 & 6.92 & 4.94 & 4.56 &4.43 &4.39 \\
        S$^2$-DMs(DDIMs) & \textbf{11.38} & \textbf{6.36} & \textbf{4.46} & \textbf{4.23} &\textbf{4.20} &\textbf{4.06} \\
        \rowcolor{orange!30}\textbf{Increase Rate} & 12.33\% & 8.09\% & 9.72\% & 7.24\% & 5.19\% & 7.52\% \\
        \noalign{\vskip 3pt}
        \hline
        \noalign{\vskip 3pt}
        PNDMs &13.67  &7.61  &4.87  &3.99  &3.67  &3.42 \\
        S$^2$-DMs(PNDMs) &\textbf{12.01}  &\textbf{6.54}  &\textbf{4.36}  &\textbf{3.77}  &\textbf{3.55}  &\textbf{3.26} \\
        \rowcolor{orange!30}\textbf{Increase Rate} & 12.14\% & 14.06\% & 10.47\% & 5.51\% & 3.27\% & 4.68\% \\
        \noalign{\vskip 3pt}
        \hline
        \noalign{\vskip 3pt}
        DEIS &4.68  &3.94  &3.77  &-  &-  &- \\
        S$^2$-DMs(DEIS) &\textbf{4.24}  &\textbf{3.78}  &\textbf{3.36}  &-  &- &-\\
        \rowcolor{orange!30}\textbf{Increase Rate} & 9.40\% & 4.06\% & 10.88\% & - & - & - \\
        \noalign{\vskip 3pt}

        \hline
    \end{tabular}
    \caption{ FID scores for the S$^2$-DMs against baseline methods trained on CelebA(64x64) with the $L_{skip}$.}
    \begin{tabular}{ccccccc}
        \hline
        \textbf{Models} \textbackslash \, \# samplesteps S & \textbf{10} & \textbf{20} & \textbf{50} & \textbf{100} & \textbf{200}  & \textbf{1000} \\
        \hline \hline
        \noalign{\vskip 3pt}
        DDIMs &13.15  &9.29  &6.40  &5.24  &4.58 &4.23 \\
        S$^2$-DMs(DDIMs) &\textbf{11.97}  &\textbf{8.12}  &\textbf{5.29}  &\textbf{4.18}  &\textbf{3.65} &\textbf{3.13} \\
        \rowcolor{orange!30}\textbf{Increase Rate} & 8.97\% & 12.59\% & 17.34\% & 20.23\% & 20.31\% & 26.00\% \\
        \noalign{\vskip 3pt}
        \hline
        \noalign{\vskip 3pt}
        PNDMs &12.59  &8.72  &6.00  &4.89  &4.30 &3.43 \\
        S$^2$-DMs(PNDMs) &\textbf{11.40}  &\textbf{7.58}   &\textbf{4.94} &\textbf{3.91}  &\textbf{3.38} &\textbf{2.94} \\
        \rowcolor{orange!30}\textbf{Increase Rate} & 9.45\% & 13.07\% & 17.67\% & 20.04\% & 21.40\% & 14.29\% \\
        \noalign{\vskip 3pt}
        \hline
        \noalign{\vskip 3pt}
        DEIS &6.93  &2.77  &2.39  &-  &- &- \\
        S$^2$-DMs(DEIS) &\textbf{6.29}  &\textbf{2.02}  &\textbf{1.77}  &-  &-  &- \\        
        \rowcolor{orange!30}\textbf{Increase Rate} & 9.24\% & 27.08\% & 25.94\% & - & - & - \\
        \noalign{\vskip 3pt}

        \hline
    \end{tabular}

    \label{table:celeba}
\end{table*}

\section{Experiments}\label{Experiments}


In this section, we demonstrate that the S\(^{2}\)-DMs outperform DDIMs \cite{song2020denoising}, PNDMs \cite{liu2022pseudo} and DEIS \cite{zhang2022fast} in image generation, achieving this with an identical number of steps and the same sampling algorithm. Moreover, the latent variables in the images generated by the S\(^{2}\)-DMs retain a high level of image features, allowing for interpolation within the latent space.

In consideration of computational resources, our experiments utilized the CIFAR10 dataset with a resolution of 32×32 and the CelebA dataset with a resolution of 64×64. For the training setup, we adopted the same architecture \cite{he2016deep}; \cite{ronneberger2015u}; \cite{kingma2014adam} as provided in the official DDIMs repository. We also ensured that all parameters were kept consistent, guaranteeing the reproducibility of our experiments. For hardware, we trained both datasets on two NVIDIA A100 GPUs, with 600K steps for CIFAR10 and 400K steps for CelebA. Model performance was evaluated based on the FID \cite{heusel2017gans}; \cite{jolicoeur2020adversarial}. Specifically, Our evaluation method followed the DDIMs repository, wherein we sampled 50,000 images and computed the FID against real images. To further guarantee reproducibility, we fixed the random seed in our experiments, ensuring that all results are replicable. (It is noteworthy that in the experiments, the models are trained based on the same parameters and methods. The results are then obtained using different sampling algorithms. Therefore, some of the results may not be consistent with those reported in original works. This is reasonable, as the parameters trained in each report are different. Our primary comparison focuses on the diffusion models trained using our method versus those trained with the original method, specifically comparing the effects of leap-step sampling under different sampling algorithms.)



In Table \ref{table:cifar10} and \ref{table:celeba}, we evaluate the quality of samples generated by models trained on the CIFAR10 and CelebA datasets, measured using the FID as the evaluation metric. We default to $skip=10$ and compare our results with DDIMs. As expected, by incorporating skip-step information, the model is able to capture a broader scope of knowledge, leading to an improved sample quality. We observed that the S$^2$-DMs consistently produces higher quality samples than original model with the same sampling steps and the same algorithm.   This demonstrates that the diffusion model enhanced by our training algorithm contributes to the improvement of sample quality.  Thus, other models only need a few lines of training code modifications to benefit from the performance boost this method offers, without any additional changes to the sampling algorithm. 

\subsection{Sample Quality and Ablation Experiment}
\begin{figure}[t]
\centering
\centering\centering
\includegraphics[width=\linewidth]{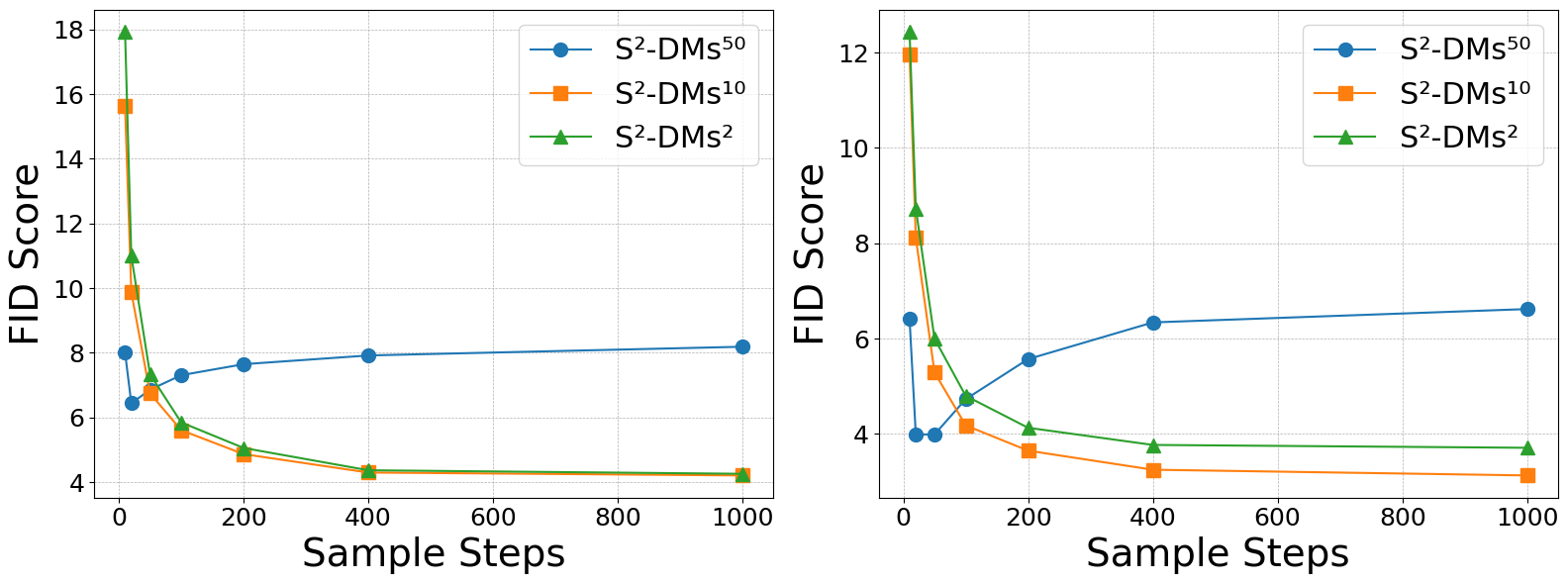}
\centering
\caption{FID scores for the step ablation on CIFAR10 and CelebA. The impact of skip steps on the model was examined by varying the skip values among \{50, 10, 2\} based on DDIMs.}
\label{fig:ablation}
\end{figure}

In Figure \ref{fig:ablation}, the impact of various skip-step intervals on the model is presented, using the same datasets (detailed further in Table \ref{tab:abl}). We employed skip-step intervals of {50, 10, 2} for the model. As expected, integrating more extensive skip-step information enables the model to obtain a wider perspective in fewer generative paths, thereby producing samples of higher quality. We believe that the introduction of $L_{skip}$ transforms the diffusion model into a skip-step diffusion model, which aligns with the number of leap steps. Models without the addition of $L_{skip}$ can be considered as having a $skip=1$ setting. Conversely, models with $skip=2$, 10, 50 settings will perform well when the sampling leap steps are 2, 10, and 50, respectively.

\begin{figure}[t]
\centering
\subfigure[Training Time and Memory]{
\begin{minipage}[t]{0.5\linewidth}
\centering\centering
\includegraphics[width=\linewidth]{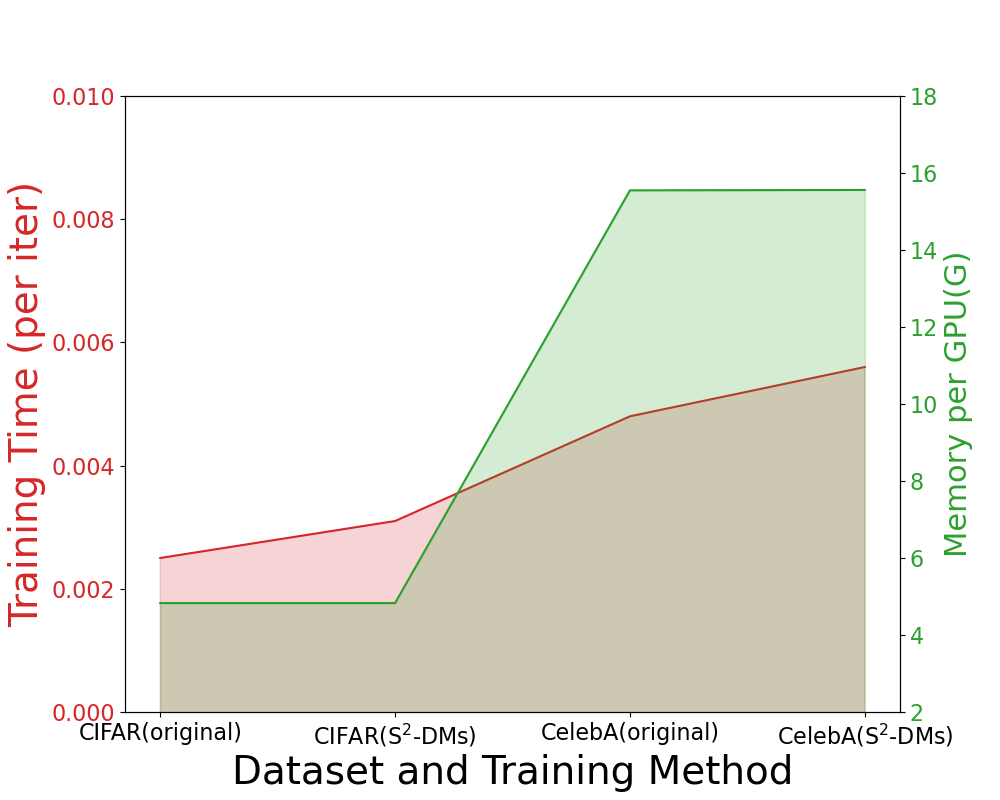}
\centering
\end{minipage}%
}%
\subfigure[Sample Time and Effects]{
\centering\begin{minipage}[t]{0.5\linewidth}
\centering
\includegraphics[width=\linewidth]{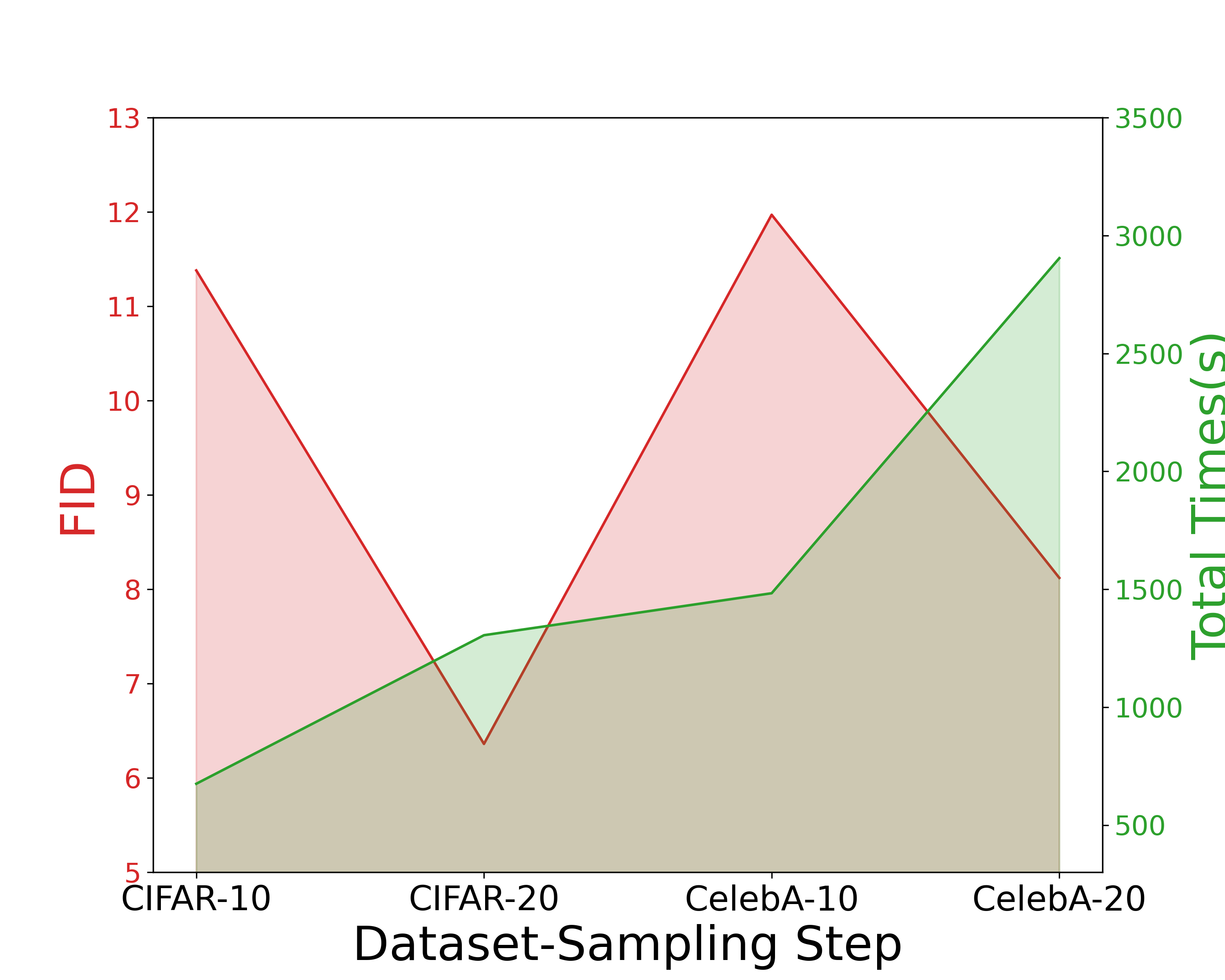}
\centering
\end{minipage}%
}%
\caption{Visualization of efficiency-effectiveness analysis.}
\label{fig:cost}
\end{figure}

In Figure \ref{fig:cost}, we specifically focused on the training time and memory usage implications of incorporating our proposed $L_{skip}$ into the model. The empirical data revealed a modest increase in training time: for instance, when integrating $L_{skip}$, the training time rose from 0.0025 seconds to 0.0031 seconds per iter on CIFAR10, and from 0.0048 seconds to 0.0056 seconds on CELEBA. Notably, this increase in training time, while measurable, remains relatively minor. Moreover, it is essential to highlight that the memory usage remained largely consistent, indicating that the incorporation of $L_{skip}$ does not exert additional pressure on memory resources. This suggests that the added computational cost is minimal and, crucially, does not compromise the efficiency of the training process. Given the negligible increase in training time and stable memory usage, we believe that the trade-off is favorable, considering the significant performance improvements our method offers.

\begin{figure*}[htb]
  \centering
  \includegraphics[width=\linewidth]{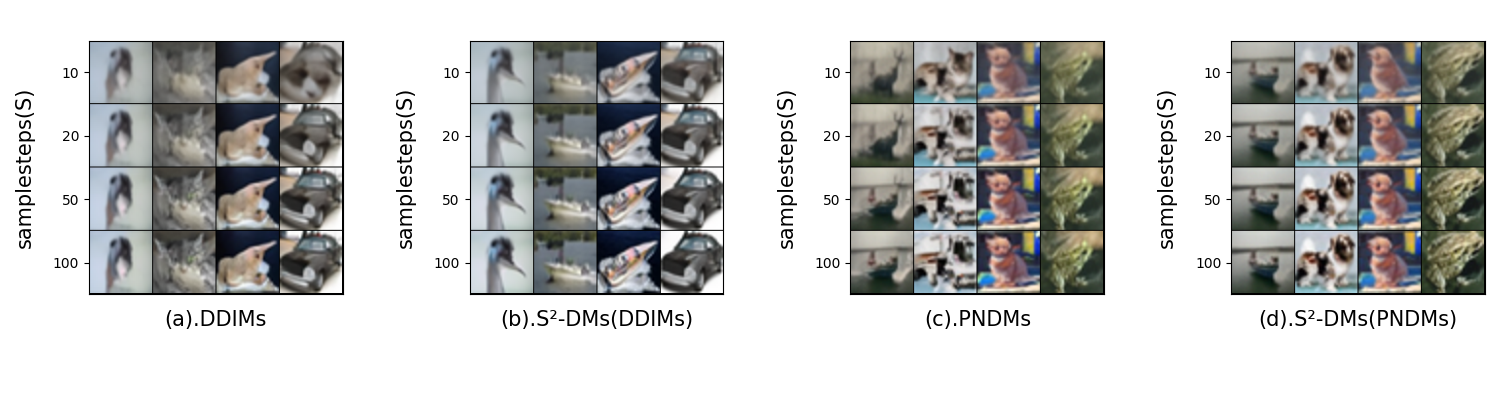}
      \vspace{-30pt}

  \caption{ CIFAR10 and CelebA samples with difference models in \{10, 20, 50, 100\} steps.}
      \label{fig:pic1}

    \centering
  \includegraphics[width=\linewidth]{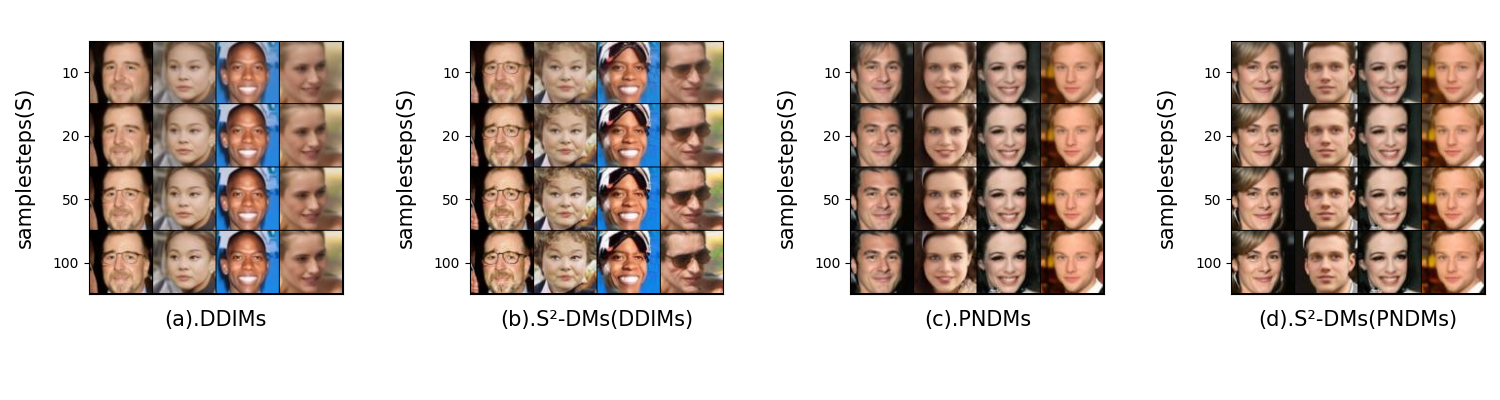}
      \vspace{-30pt}

  \caption{ CIFAR10 and CelebA samples with difference models in \{10, 20, 50, 100\} steps.}
    \label{fig:pic2}
\end{figure*}

In Figures \ref{fig:pic1} and \ref{fig:pic2}, we display samples from the CIFAR10 and CelebA datasets generated by models with identical numbers of sampling steps and the same sampling algorithm. These figures reveal that, under the same sampling conditions, the samples generated by S$^2$-DMs are of higher quality. Additionally, as the number of sampling steps increases, so does the quality of the samples. For instance, in CIFAR10, the images of boats and cars in the second and fourth columns demonstrate that S$^2$-DMs can produce highly accurate images in just 10 steps, whereas DDIMs yield blurrier images and require more steps for high-quality outcomes. Similarly, in CelebA, the third column presents images of a male subject, where S$^2$-DMs generate a clear hat, in contrast to the blurrier hat produced by DDIMs. Moreover, the detail in images generated by PNDMs is less pronounced compared to those by S$^2$-DMs. This difference is significant, highlighting the substantial improvement in sample quality achieved by S$^2$-DMs with the same number of sampling steps. 
\subsection{Interpolation and Generation Consistency}


The S$^2$-DMs, drawing from the deterministic generation frameworks of DDIMs \cite{song2020denoising}, PNDMs \cite{liu2022pseudo}, and DEIS \cite{zhang2022fast}, also incorporate the semantic interpolation typical of implicit models like GANs \cite{mohamed2016learning}. This dual approach allows S$^2$-DMs to blend structured generation with creative interpolation capabilities.

\begin{figure}[htb]
  \centering
  \includegraphics[width=\linewidth]{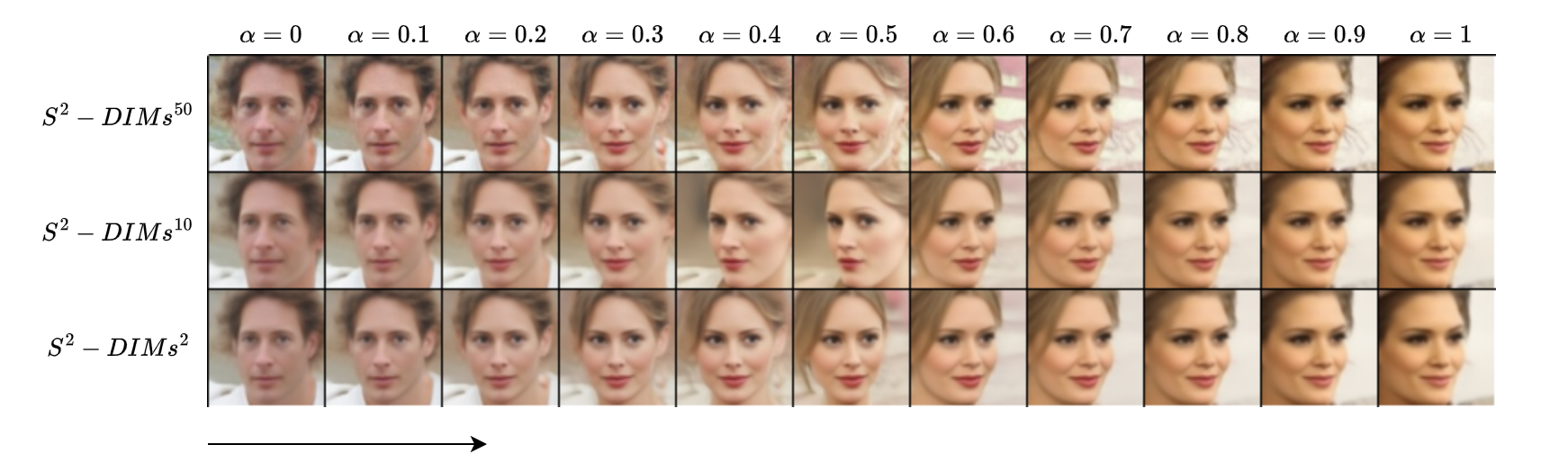}
  \caption{Interpolation of samples from the S$^2$-DMs in 10 steps. The $\alpha$ represents the weight used for interpolation, moving from left to right.}
 \label{inter}
\end{figure}

In Figure \ref{inter}, the interpolation outcomes of S$^2$-DMs under various skip-step settings are displayed. The results indicate that even basic interpolation in $x_T$ achieves semantically rich transitions between two samples. Particularly at a 'skip' of 50, the models generate samples with remarkable quality and detailed features. This is evidenced in images 5 to 7, where the models adeptly reproduce light and shadow effects on facial features, demonstrating the nuanced capabilities of S$^2$-DMs.

Conversely, at a $skip$ of 2, the model's behavior is more akin to the original DDIMs, leading to slightly less detailed samples. This highlights the impact of skip-step intervals on the fidelity of generated images.

Moreover, the figure shows that S$^2$-DMs maintain a consistent level of quality across different training settings, even when conditioned on the same $x_T$ encoding. This underscores the robustness and adaptability of S$^2$-DMs, capable of producing stable and high-quality outputs across various skip-step configurations.


\section{Related Work}
Denoising Diffusion Probabilistic Models (DDPMs; \cite{ho2020denoising}) and Noise Conditional Score Networks (NCSNs; \cite{song2019generative}) are notable for their sample quality, comparable to GANs. While DDPMs optimize a variational lower bound, NCSNs target score matching over a Parzen density estimator. Both models employ a denoising autoencoder across noise levels and use Langevin dynamics for sampling. The shared approach requires multiple iterations for optimal sample quality. Recent advancements aim to decrease DDPMs' inference steps through SDE solvers and programming algorithms. However, challenges like the disparity between log-likelihood reduction and FID \cite{heusel2017gans}, \cite{szegedy2016rethinking} remain in some models.

DDIMs \cite{song2020denoising} emerges as an implicit generative model, wherein samples are uniquely defined by latent variables. This lends DDIMs properties akin to GANs and invertible flows, including the capability to produce semantically meaningful interpolations. Conceived from a purely variational standpoint, DDIMs sidesteps the constraints of Langevin dynamics, potentially explaining its superior sample quality compared to DDPMs in fewer iterations. The sampling paradigm of DDIMs also echoes the concepts found in neural networks with continuous depth. Additionally, other innovative methods have been introduced to further refine DDPMs sampling, such as reverse SDEs \cite{song2020score} with unique coefficients, ``corrector" steps, and probability flow ODEs \cite{liu2022flow}. As the exploration of efficient sampling in diffusion models continues, our research stands on the shoulders of these pioneering works, aiming to push the boundaries of what's achievable in generative models.

\section{Conclusion and Future Work} \label{future}


We propose the Skip-Step Diffusion Models, a diffusion model that adapts better to accelerated sampling algorithms by simply adding an additional leap step loss during the training process. We demonstrate how to incorporate leap step loss into the original loss function during training. Our results qualitatively and quantitatively show that under the same accelerated sampling algorithms, S$^2$-DMs significantly improve the sample quality of image generation. Our method successfully explores adding leap step information during training, allowing the model to reach a symmetrical state during sampling, thereby enhancing sample quality.

Our findings pave a new direction for future diffusion model research. By investigating the asymmetry between the training of diffusion models and accelerated sampling algorithms, we can ensure that the trained models are better suited to accelerated sampling algorithms. This ensures that even with algorithms using fewer sampling steps, the model can still generate high-quality outputs. Perhaps this offers an effective solution to the challenges diffusion models face in balancing sampling speed and sample quality. In the future, we will continue to explore how to better integrate leap-step information to ensure greater consistency between the trained model and accelerated sampling algorithms, thereby improving sample quality. Additionally, we plan to investigate how to incorporate leap-step information into ODEs, while also exploring possibilities in non-continuous spaces. 
\bibliography{example_paper}
\bibliographystyle{icml2024}
\newpage
\onecolumn
\appendix

\section{Experimenal Details}
In this section, we include more details about the training and sampling of the S$^2$-DMs. All the experiments are run on two NVIDIA A100 GPUs.
\subsection{Training and Cost}
Our experiments were conducted on the CIFAR10 and CelebA. Since the original size of the CelebA dataset is not 64x64, we followed Song's approach by first center-cropping the CelebA images and then resizing them to 64x64 dimensions. During model training, we set the batch size to 128 and employed the Adam optimizer. On the CIFAR10 and CelebA datasets, we iterated for 600K and 400K times respectively, even though the typical number of iterations stated is 800K/600K. To investigate this effect, the main text also showcases experimental results from training for 800K/600K iterations. Finally, the images generated by the model were compared with a dataset of 50,000 images for FID calculation.

\begin{table}[h]
    \caption{ Training time (s) per iteration on the S$^2$-DMs.}
    \centering    
    \begin{tabular}[]{ccccc}
        \noalign{\vskip 3pt}
        \hline
        \textbf{Dataset/Model} &CIFAR/DDIMs &CelebA/DDIMs &CIFAR/S$^2$-DMs &CelebA/S$^2$-DMs \\
        \noalign{\vskip 3pt}

        \hline \hline
        \noalign{\vskip 3pt}
        Time per iter(s)  &0.0025 &0.0048 &0.0031  &0.0056 \\
         Memory per GPU(G)&4.83 &15.55 &4.83  & 15.56\\
        \noalign{\vskip 3pt}
         \hline
    \end{tabular}
    \label{tab:time1}
\end{table}

\begin{table}[h]
    \caption{ Sample total times (s) on the S$^2$-DMs. Sampling 50K samples, each iteration is 10K/2K in CIFAR10/CelebA.}
    \centering    
    \begin{tabular}{ccccc}
        \hline
        \noalign{\vskip 3pt}
        \textbf{Datasets-samplesteps} &CIFAR-10 &CIFAR-20 &CelebA-10 &CelebA-20 \\
        \noalign{\vskip 3pt}
        \hline \hline
        \noalign{\vskip 3pt}
        Total Times(s)  &675 &1305 &1483  &2904 \\
        \noalign{\vskip 3pt}
         \hline
    \end{tabular}
    \label{tab:time2}
\end{table}
To ensure the repeatability of our experiments, we uniformly adopted the random seed 1234 from Song's repository by default. Additionally, on the CIFAR10 dataset, we refrained from utilizing multi-threading, guaranteeing that the reproducibility of the experiments would not be compromised by hardware randomness.

\begin{table*}[th]
\caption{ FID scores for the stpe ablation on CIFAR10 and CelebA. The impact of skip steps on the model was examined by varying the skip values among \{50, 10, 2\} based on DDIMs.  }
\label{tab:abl}
\begin{center}
\begin{tabular}{cccccccc}
\hline \\[-0.7em]
\textbf{Models} \textbackslash \, \# samplesteps S & \textbf{10} & \textbf{20} & \textbf{50} & \textbf{100} & \textbf{200} & \textbf{400} & \textbf{1000} \\ \\[-0.7em]
\hline \\[-0.5em]
CIFAR10(32×32) & & & & & & & \\ \\[-0.5em] 
\hline \\[-0.5em]
S$^2$-DMs$^{50}$ & \textbf{8.01} & \textbf{6.44} & 6.86 & 7.31 &7.65 &7.92 &8.19 \\
S$^2$-DMs$^{10}$ & 15.63 & 9.88 & \textbf{6.75} & \textbf{5.61} &\textbf{4.87}  &\textbf{4.30} &\textbf{4.21} \\
S$^2$-DMs$^{2}$ & 17.92 & 11.00 & 7.34 & 5.85 &5.06  &4.37 &4.26 \\

\\[-0.5em]
\hline \\[-0.5em]
Celeba(64×64) & & & & & & & \\ \\[-0.5em]
\hline \\[-0.5em]
S$^2$-DMs$^{50}$ &\textbf{6.41}  &\textbf{3.99} &\textbf{3.99}  &4.73  &5.57  &6.34 &6.62 \\ 
S$^2$-DMs$^{10}$ &11.97  &8.12  &5.29  &\textbf{4.18}  &\textbf{3.65}  &\textbf{3.25} &\textbf{3.13} \\
S$^2$-DMs$^{2}$ &12.43  &8.73  &6.00  &4.80  &4.13  &3.77 &3.71 \\\\[-0.5em]
\hline \\[-0.5em]
\end{tabular}
\end{center}
\end{table*}

We quantitatively investigated the training overhead of the model. All results were measured on two NVIDIA A100 graphics processors. In Table \ref{tab:time1}, we report the time consumption and memory usage for each iteration on the CIFAR10 and CelebA datasets. As can be observed, the introduction of $L_{skip}$
  increased the training overhead of the model. However, it did not significantly extend the overall training time. We believe that compared to the substantial performance improvement, the added overhead is acceptable.

  \subsection{Sampling}
In Table \ref{tab:time2}, we showcase the time required for the S$^2$-DMs to sample with \{10, 20\} steps on different datasets. We believe that within this range, the trade-off between performance and time cost is optimal.

It's worth noting that when we trained and sampled the original DDIMs model on CelebA using mixed-precision training, we encountered issues related to gradient explosion. However, this problem did not arise when employing the same mixed-precision training with the S$^2$-DMs. We plan to investigate this issue further. For now, we believe it leans more towards engineering and hardware-related challenges.

In Table \ref{tab:abl}, we provide detailed numerical results of ablation experiments with different values of $skip$. The best results are highlighted in bold. From the table data, it's evident that with smaller sample steps, the larger the $skip$, the better the model performs, as it gets closer to the symmetric interval at this point.

\end{document}